\documentclass{ecai}

\usepackage{times}
\usepackage{latexsym}
\usepackage{booktabs}
\usepackage{helvet}
\usepackage{courier}
\usepackage{graphicx}
\usepackage{amsmath,amsfonts,bm,amsthm,mathtools}

\usepackage{algorithmic}
\usepackage{algorithm}

\usepackage[T1]{fontenc}

\usepackage[utf8]{inputenc}

\usepackage{microtype}

\usepackage{inconsolata}

\begin{document}

\begin{frontmatter}

\title{Efficient GPT Model Pre-training using \\ Tensor Train Matrix Representation }

\author{Viktoriia Chekalina\textsuperscript{\rm 1}, Georgii Novikov\textsuperscript{\rm 1}, Julia Gusak\textsuperscript{\rm 1}, Ivan Oseledets\textsuperscript{\rm 1} \textbf{and Alexander Panchenko}\textsuperscript{\rm 1} \\
  \textsuperscript{\rm 1}Skolkovo Institute of Science and Technology} 



\begin{abstract}
Large-scale transformer models have shown remarkable performance in language modelling tasks. However, such models feature billions of  parameters, leading to difficulties in their deployment and prohibitive training costs from scratch. To reduce the number of the parameters in the GPT-2~\cite{GPT_2} architecture, we replace the matrices of fully-connected layers with the corresponding Tensor Train Matrix~(TTM)~\cite{oseledetsTTM} structure. Finally, we customize forward and backward operations through the TTM-based layer for simplicity and the stableness of further training.
The resulting GPT-2-based model stores up to 40\% fewer parameters, showing the perplexity comparable to the original model. 
On the downstream tasks, including language understanding and text summarization, the model performs similarly to the original GPT-2 model. The proposed tensorized layers could be used to efficiently pre-training other Transformer models. 
\end{abstract}

\end{frontmatter}

\section{Introduction}

Large language models such as GPT-2, GPT-3~\cite{GPT_2, GPT_3} show outstanding results in all areas of natural language processing. However, training and employing models with a vast number of parameters needs memory, time and electricity proportional to model size. 

To make GPT-2-based models easier to deploy, we replaced fully connected layers with sequential TTM~\cite{oseledetsTTM} containers, based on Tensor Train~(TT)~\cite{Oseledets11} representation. We tested several approaches to forward and back propagations of a signal through containers and chose the most memory-stable and time-optimal pattern. The weight matrix is generally full-rank and can't be approximated with low-rank objects. Therefore, we trained the architecture with custom TTM layers from scratch: thus, we were looking for the weights of the linear layer not among all matrices but among those represented in the TTM format. Then we study the behaviour of the pre-trained custom model on in-domain and out-off-domain language modelling tasks and several downstream tasks.

The contribution of our paper is the following:
%
(i) We develop a custom TT-layer that, firstly, has fewer parameters and, secondly, uses less memory during forward and backward passes;
(ii) We provide a GPT-based model with up to 40\% fewer parameters showing performance close to the original GPT in in-domain and out-of-domain tasks language modelling, GLUE benchmark, and text summarization.

\begin{figure*}[h]
\begin{center}
  \includegraphics[width=0.9\linewidth]{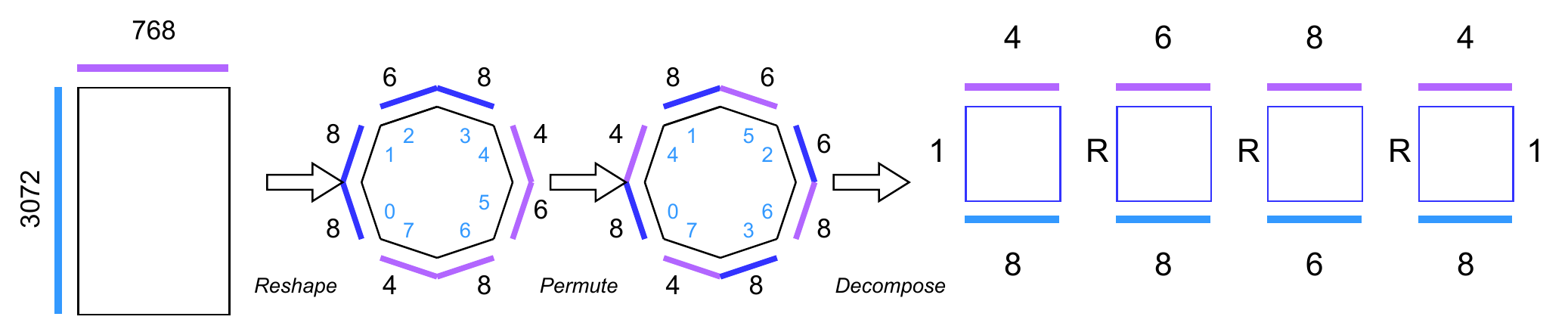}
  \caption{The scheme of TTM representation of linear layer in a GPT-2 small MLP block. Black digits indicate the size of the axes, and light blue - their number.}
  \label{ttm_scheme}
\end{center}
\end{figure*}

\section{Related work}\label{sec:related_work}

Several approaches explore ways of reduction of the size of language models. The mechanism of distillation~\cite{distillation0} was applied to BERT~\cite{distill_bert} and GPT-2\footnote{https://huggingface.co/distilgpt2}. The Open
Pre-trained Transformers (OPT)~\cite{Opt} provide a smaller model that imitates the behaviour of GPT-3~\cite{GPT_3}. They employ more efficient training and use the particular datasets for improving generalization capability.

TT~(Tensor Train) is an effective way to obtain low-rank representaions of inner layers and is also used to reduce parameter numbers. \cite{Khrulkov} and~\cite{facebook_tt} reduce the size of the embedding layer using TT. \cite{NIPS2015_6855456e} uses the TT format of linear layers to compress the computer vision models, however TT representations were not tested before for generative Transformers.


\begin{table}
\centering
\scalebox{0.97}{

\begin{tabular}{@{}l| cccc }
\toprule
Layer &  TTM-16 & TTM-32 & TTM-64 & Fully-Connected \\
Type &   &  &  &  \\
\midrule
Memory, GB & 48.7 & 75.07  & 48.31 & 48.37 \\
\bottomrule
\end{tabular}
}
    
\caption{Peak memory footprints for signal propagation in full GPT-2 model with TTM layers with different ranks. At the rank 16 we have an increment in memory consumption.}
\label{tab:model_peak_memory_old_new}
\end{table}

\begin{table}
\centering
\scalebox{0.97}{

\begin{tabular}{@{}l| ccc }
\toprule
Layer &  TTM-16 & TTM-16 & Fully-Connected \\
\midrule
Backprop Strategy & PyTorch & Einsum  & PyTorch \\
     & Autodiff & Full Matrix  &  Autodiff\\
\midrule 
\addlinespace
Single Layer, Batch 16 & 1100 MB & 294 Mb & 395 Mb \\
\bottomrule
\end{tabular}
}
    
\caption{Memory footprints for signal propagation in TTM wiht rank 16 and Fully-Connected Layers. PyTorch strategy leads to memory costs for TTM.}
\label{tab:ttm_layer_memory_old_new}
\end{table}

\begin{table}
\fontsize{9pt}{9.5pt}\selectfont%
\centering
\begin{tabular}{@{}llrr@{}}
\toprule
Forward & Backward & Memory, Mb & Time, ms \\
\midrule

Einsum & PyTorch Autodiff & 1008 & 23.6 \\
Einsum  & Full Einsum  & 192 & 55.7 \\
\textbf{Einsum}  & \textbf{Full Matrix}  & \textbf{192} & \textbf{17.5} \\
Fixed Scheduler & PyTorch Autodiff & 2544 & 58.4 \\ 
Fixed Scheduler & Full Einsum  & 192 & 84 \\
Fixed Scheduler  & Full Matrix  & 192 & 125 \\
\bottomrule
\end{tabular}
\caption{Time and memory footprints for different forward and backward strategies for TTM-16 layer.}
\label{tab:forward_backward}
\end{table}

\section{Math Background}

{\bf Notation.} We denote vectors as $\mathbf{v}$, matrices as $\mathbf{M}$ and tensors of 3-rd order and higher as $\mathcal{T}$.

{\bf Tensor contraction.} Given two tensors ${\mathcal{T}^1 \in \mathbb{R}^{I_1\times\dots\times I_M\times S_1\times\dots\times S_K}}$ and ${\mathcal{T}^2 \in \mathbb{R}^{S_1\times\dots\times S_K\times J_1\times\dots\times J_N}}$ the result of {\it tensor contraction} along axis $s_1,\dots, s_K$ is a tensor $\mathcal{T}\in\mathbb{R}^{I_1\times\dots\times I_M\times J_1\times\dots\times J_M}$, where one element is computed using formula 

$$\mathcal{T}_{i_1,\dots, i_M, j_1,\dots,j_N} =$$
$$= \sum\limits_{s_1,\dots,s_K}\mathcal{T}^1_{i_1,\dots,i_M, s_1,\dots, s_K}\mathcal{T}^2_{s_1,\dots, s_K,j_1,\dots,j_N}$$

and requires ${O(S_1S_2\dots S_K)=O\left(\prod\limits_{k=1}^K S_k\right)}$
floating point operations (FLOP). Thus, number of FLOP to compute tensor $\mathcal{T}$ is $O\left(\prod\limits_{m=1}^M I_m \prod\limits_{n=1}^N J_n \prod\limits_{k=1}^K S_k\right)$.
For example, a multiplication of two matrices of shapes $(I, S)$ and $(S, J)$ can be calculated for $O(IJS)$ operations.

{\bf TTM format.} We say that a tensor $\mathcal{T} \in \mathbb{R}^{I_1\times J_1\times \dots\times I_M \times J_M}$ is represented in {\it Tensor Train Matrix (TTM) format} with rank $(R_0, R_1,\dots, R_M)$ if each element is computed as 
$$\mathcal{T}_{i_1,j_1,\dots,i_M,j_M} = $$
$$\sum\limits_{r_1,\dots, r_{M-1}} \mathcal{G}^1_{r_0, i_1, j_1, r_1}\dots \mathcal{G}^M_{r_{M-1}, i_M, j_M, r_M},$$
where ${\mathcal{G}^m\in\mathbb{R}^{R_{m-1}\times I_m\times J_m\times R_m}}$, $m=\overline{1,{M-1}}$ are {\it core tensors (cores)} of TTM decomposition. Note that $R_0=R_M=1$.

Assume that $R_m = R$ for $m=\overline{1,{M-1}}$, then to represent tensor $\mathcal{T}$ with $\prod_{m=1}^M I_mJ_m$ elements we need to store only $R(I_1J_1 + I_MJ_M) + R^2\sum_{m=2}^{M-1}I_mJ_m$ parameters of core tensors.

\textbf{Forward pass in TTM as a contraction process}
We represent $\mathcal{W_M}$ as a set of $M$ cores $\mathcal{G}$, so every element in $\mathcal{W_M}$ is enumerated by a 2M-tuple of indices and is defined as:
$$\mathcal{W}_{(i_1, j_1),(i_2, j_2), . . . ,(i_M, j_M)} = $$
$$ \mathcal{G}^1(:, i_1, j_1, :)\mathcal{G}^2(:, i_2, j_2, :). . . \mathcal{G^M}(:, i_M, j_M, :)$$
So, in equation~\ref{eq:ttm_forward_pass} we should contract X with sequence ($\mathcal{G}^M$, …, $\mathcal{G}^1$) sequentially. Please note, we can start with the first core $(\mathcal{G}^1,...\mathcal{G}^M$) or with the last ($\mathcal{G}^M$, …, $\mathcal{G}^1)$, in common it doesn’t matter.

We contract $\mathcal{X}$ with size $(B, D_in)$ to $G_M$ with size $(R_{M-1}, I_M, J_M, 1)$. As $D_in = \prod (I_1 … I_M)$, we contract over $I_M$ and have a tensor of shapes
$(B, R_{M-1}, J_M, I_{M-1}, …, I_1)$ as a result. 
This tensor we should contract to core $\mathcal{G}_{M-1}$ with shapes 
$$(R_{M-2}, I_{M-1}, J_{M-1}, R_{M-1})$$
over $I_{M-1} R_{M-1}$ dimensions. This operation yields the object of shapes $$(B, R_{M-2}, J_M, J_{M-1}, I_{M-2}, \dots, I_1)$$.

By repeating such operation K times, we obtain product with shapes 

$$(B,I_1,...,I_K,J_{K+1},...,J_M,R_K)$$.
In the end, we gain the output of sizes $(B, J_1, …J_M) = (B, D_{out})$.


The computational complexity of this operation is estimated above.

\section{Efficient TTM Layer}
As we describe in Section~\ref{sec:related_work}, replacing linear layers with layers whose weights are represented in a factorized format allows for reducing memory and time complexity of the network. 
In our paper we focus on replacing linear layers with TTM layers. In TTM layer the weight $\mathbf{W}$ is a matrix of shape ${D_{in}\times D_{out}}$ represented in TTM format with $M$ cores ${\mathcal{G}^m\in\mathbb{R}^{R_{m-1}\times I_m\times J_m\times R_m}}$, $m=\overline{1,{M}}$, where $I_m$ and $J_m$ are such that ${D_{in} = \prod\limits_{m=1}^MI_m}$ and ${D_{out} = \prod\limits_{m=1}^MJ_m}$.
%
%
We call $(R_1,\dots,R_{M-1})$ a rank of TTM. For example, a Fully-Connected Transformer layer of size $[768\times 3072]$ is represented as sequence of a TTM-R cores of sizes $[1, 8, 4, R]$, $[R, 8, 6, R]$,  $[R, 6, 8, R]$, $[R, 8, 4, 1]$, as it depicted in Fig.~\ref{ttm_scheme}. Dimensions marked by blue correspond to the input in the initial matrix, marked by violet - by output. The forward signal propagation means sequentially contraction of input vector $X$ with core tensors. We notice that the TTM layer and models with these layers might allocate more or less memory during training than the classical ones depending on the rank. 

We measure the peak memory during one training iteration in GPT-2 model with TTM layers with different ranks~ (Table~\ref{tab:model_peak_memory_old_new}) and observe a significant increase in memory consumption at rank 16. The memory footprint for Fully-Connected and TTM layers with this rank custom and PyTorch signal propagation strategies~(Table~\ref{tab:ttm_layer_memory_old_new}) confirms it. 
 We extend the existing research by proposing memory-efficient techniques to compute forward and backwards through the TTM layer for a more comprehensive description of the proposed methods.

\subsection{Forward pass}
{\bf Fully-connected layer}.
Given an input batch ${\mathbf{X}\in\mathbb{R}^{B\times D_{in}}}$ a forward pass through a fully-connected layer with weight matrix ${\mathbf{W} \in \mathbb{R}^{D_{in} \times D_{out}}}$ and bias vector $\mathbf{b}\in\mathrm{R}^{D_{out}}$ results in the output $\mathbf{Y} = \mathbf{X}\mathbf{W} + \mathbf{b} \in \mathbb{R}^{B\times D_{out}}$ and requires $O(BD_{in}D_{out})$ operations.

{\bf TTM layer}. 
The weight of TTM layer is represented with cores
$\mathcal{G}^m, m=\overline{1,M}$.
If we sequentially contract input $\mathcal{X}$ with cores, then after contracting with $\mathcal{G}^M,\dots,\mathcal{G}^{k+1}$ we get a tensor of shape $(B, I_1,\dots, I_K, J_{K+1},\dots, J_M, R_{K})$, its contraction with the next core $\mathcal{G}^k \in\mathbb{R}^{R_{k-1}\times I_k\times J_k\times R_k}$ requires~$BI_1\dots I_k J_k\dots J_{M}R_{k-1}R_k$ steps. Thus the computational complexity of the TTM layer is $O(BM\max\{D_{in}, D_{out}\}\max\limits_k\{I_k, J_k\}(\max\limits_kR_k)^2)$ and depends on the schedule in which we contract cores.

{\bf TTM layer: Einsum}.
The schedule of contractions computed during the forward pass is optimized via $opt\_einsum$ function. Thus, due to some shared intermediate results, memory for saved activations might be optimized~(see Algorithm~\ref{alg::forward_ttm_einsum}). 
\begin{algorithm}[H]
\begin{algorithmic}[1]
\REQUIRE data $\mathbf{X}\in\mathbb{R}^{B\times D_{in}}$;
parameters ${\mathbf{W}\in\mathrm{R}^{D_{in}\times D_{out}}}$, ${\mathbf{b}\in\mathrm{R}^{D_{out}}}$;
\ENSURE $\mathbf{Y} = \mathbf{X}\mathbf{W} + \mathbf{b}\in\mathrm{R}^{B\times D_{out}}$
\end{algorithmic}
 \caption{Forward pass (Fully-connected layer). Number of layer parameters is $O(BD_{in}D_{out})$. Computational complexity is $O(BD_{in}D_{out})$. SavedActivations is $O(BD_{in})$.}
 \label{alg::forward_fc}
\end{algorithm}

{\bf TTM layer: Fixed Scheduler}.
The order of cores to contract with is fixed in advance, and we don't optimize it with $opt\_einsum$. In this case, saved activations usually occupy the same amount of memory~(see Algorithm~\ref{alg::forward_ttm_byhands}).

\begin{algorithm}[H]
\begin{algorithmic}[1]
\REQUIRE 
${\text{data } \mathbf{X}\in\mathbb{R}^{B\times D_{in}}}$; 
${D_{in} = \prod\limits_{m=1}^MI_m, D_{out} = \prod\limits_{m=1}^MJ_m}$;
${\text{parameters } \mathcal{G}^m\in\mathrm{R}^{R_{m-1}\times I_m\times J_m\times R_m}, m=\overline{1,M}}$, ${R_0=R_M=1}$;
\ENSURE $\mathcal{Y} \in\mathrm{R}^{B\times J_1\times\dots\times J_M}$
\STATE $\mathcal{X} = Reshape(\mathbf{X}) \in\mathrm{R}^{B\times I_1\times\dots\times I_M}$
\STATE $\mathcal{Y} = einsum(\mathcal{G}^1,\dots\mathcal{G}^M, \mathcal{Y})$
\end{algorithmic}
 \caption{Forward pass (TTM layer, Einsum). Number of layer parameters is $O(\sum\limits_{m=1}^M R_{m-1}I_mJ_mR_m)$.}
 \label{alg::forward_ttm_einsum}
\end{algorithm}

\begin{algorithm}[H]
\begin{algorithmic}[1]
\REQUIRE 
${\text{data } \mathbf{X}\in\mathbb{R}^{B\times D_{in}}}$; 
${D_{in} = \prod\limits_{m=1}^MI_m, D_{out} = \prod\limits_{m=1}^MJ_m}$;
${\text{parameters } \mathcal{G}^m\in\mathrm{R}^{R_{m-1}\times I_m\times J_m\times R_m}, m=\overline{1,M}}$, ${R_0=R_M=1}$;
\ENSURE $\mathcal{Y} \in\mathrm{R}^{B\times J_1\times\dots\times J_M}$
\STATE $\mathcal{X} = Reshape(\mathbf{X}) \in\mathrm{R}^{B\times I_1\times\dots\times I_M}$
\STATE $\mathcal{Y}_0:=\mathcal{X}$
\STATE $ContractionSchedule := (1, 2,\dots,M)$
\FOR{$k$ in $ContractionSchedule$}
\STATE $\mathcal{Y}_{k} = einsum(\mathcal{G}^k, \mathcal{Y}_{k-1})$
\STATE ${//FLOP_{\mathcal{Y}} = O(B\prod\limits_{m=1}^{k+1}J_m\prod\limits_{m=k+1}^{M}I_mR_kR_{k+1})}$
\STATE ${//Memory_{\mathcal{Y}} = O(B\prod\limits_{m=1}^{k}J_m\prod\limits_{m=k+1}^{M}I_mR_k)}$
$\mathcal{Y}=\mathcal{Y}_{k}$
\ENDFOR
\end{algorithmic}
 \caption{Forward pass (TTM layer, Fixed Scheduler). Number of layer parameters is $O(\sum\limits_{m=1}^M R_{m-1}I_mJ_mR_m)$.}
 \label{alg::forward_ttm_byhands}
\end{algorithm}




\begin{table*}[h!]
\centering
\scalebox{0.95}{
\begin{tabular}{l | c | c | c | c | c}
\toprule
Model & Training & Validation & Number of & \% of classic & Perplexity\\
 &  &  & parameters & GPT-2 size &\\
\midrule
GPT-2 small & Wikitext-103 train &  Wikitext-103 test & 124439808 & 100 & 17.55 \\ 
GPT-2 small TTM-16 & Wikitext-103 train &  Wikitext-103 test& 68085504  & 54 & 21.33 \\
GPT-2 small TTM-32 &  Wikitext-103 train &  Wikitext-103 test & 71756544 & 57 & 21.06  \\
GPT-2 small TTM-64 &  Wikitext-103 train &  Wikitext-103 test & 83606784 & 67 & 18.08 \\
GPT-2 small TTM-80 &  Wikitext-103 train &  Wikitext-103 test& 107698944& 86 & 17.61 \\
\bottomrule
\end{tabular}}
\caption{In-domain perplexities for GPT-2 small model, pre-training from scratch.}
\label{table_gpt_small_en}
\end{table*}

\begin{table*}
\centering
\scalebox{0.95}{
\begin{tabular}{l | c | c | c | c | c | c}
\toprule
Model & Training & Validation & Fine-tune & Number of & \% of classic & Perplexity\\
 &  &  & & parameters & GPT-2 size &\\
\midrule
GPT-2 med & Webtext &  Wikitext-103 & No & 354823168 & 100 & 20.56 \\ 
GPT-2 TTM-72 & Openwebtext &  Wikitext-103 & No & 218303488  & 61 & 30.85 \\
GPT-2 SVD-50 & Openwebtext &  Wikitext-103 & No & 220920832 & 62 & 55.46 \\ 
Distil GPT-2 & Openwebtext &  Wikitext-103 & No & 81912576 & 23 & 51.45 \\ 
OPT 350m & Openwebtext + BookCorpus &  Wikitext-103& No & 331196416  & 93 & 24.75 \\
 & + Pile~\cite{Pile} & & &  & &  \\

\bottomrule
\end{tabular}}
\caption{Out-domain perplexities for GPT-2 Medium and GPT TTM-72 models, pre-training from scratch.}
\label{table_gpt_medium_en}
\end{table*}

\subsection{Backward pass}
While training neural networks, intermediate activations are saved during the forward pass to compute gradients during the backward pass. 

{\bf TTM layer: Autodiff.} Using automatic Pytorch differentiation during backpropagation through the TTM layer results in storing many intermediate activations, as the TTM layer is considered as a sequence of linear layers.
We propose several ways to perform backward pass that require minor memory consumption. 

\begin{algorithm}[H]
\begin{algorithmic}[1]
\REQUIRE $\frac{\partial L}{\partial\mathcal{Y}}$; $\mathcal{X}$ 
\ENSURE $\frac{\partial L}{\partial\mathcal{X}}, \frac{\partial L}{\partial\mathcal{G}^1},\dots,\frac{\partial L}{\partial\mathcal{G}^M}$

\STATE $\frac{\partial L}{\partial\mathbf{W}} = einsum(\frac{\partial L}{\partial\mathcal{Y}}, \mathcal{X})$ // $einsum$ here contracts only along batch dimension
\STATE //$FLOP_{\frac{\partial L}{\partial\mathbf{W}}} = O(BD_{in}D_{out})$
\STATE //$Memory_{\frac{\partial L}{\partial\mathbf{W}}} = O(D_{in}D_{out})$
\STATE // Results in the below for-cycle are computed only for the first batch during training and reused for others.
\FOR{$k$ in $\{1,\dots,M\}$} 
\STATE Compose $einsum_k$ expression for $\frac{\partial L}{\partial\mathcal{G}^k}$
\STATE Optimize contraction schedule for composed $einsum_k$
\ENDFOR
\FOR{$k$ in $\{1,\dots,M\}$} 
\STATE $\frac{\partial L}{\partial\mathcal{G}^k} = einsum_k(\frac{\partial L}{\partial\mathbf{W}}, \mathcal{G}^1,\dots,\mathcal{G}^M)$
\STATE //$FLOP_{\frac{\partial L}{\partial\mathbf{G}^k}} = O(D_{in}D_{out}\max\limits_m(I_m,J_m)(\max\limits_mR^m)^2)$
\ENDFOR
\end{algorithmic}
 \caption{Backward pass (TTM layer, Full Matrix).}
 \label{alg::backward_full_matrix}
\end{algorithm}

{\bf TTM layer: Full Einsum.}
In the first approach for each core tensor $\mathcal{G}_m$ we compute the loss gradient with respect to its parameters: 
$$\frac{\partial \mathcal{L}}{\partial \mathcal{G}_m} = \frac{\partial \mathcal{L}}{\partial \mathbf{W}} \frac{\partial\mathbf{W}}{\partial \mathcal{G}_m} = \mathbf{X}^T\frac{\partial \mathcal{L}}{\partial \mathcal{Y}} \frac{\partial\mathbf{W}}{\partial \mathcal{G}_m}.$$
As a gradient computation might be considered as a tensor contraction along a specified axis, the process includes three main steps. Firstly, generate a string-type expression, which defines the shapes of input and resulting tensors (e.g. "ikl,lkj->ij" for performing tensor contraction along two axes. Secondly, contraction schedule is defined (e.g., firstly along axis 'l' and then along axis 'k'). And thirdly, einsum computation is performed. First two steps should be done only for the first batch during training and performed independently for all $\frac{\partial \mathcal{L}}{\partial \mathcal{G}_m}$. The third step, in turn, tracks simultanuosly what contractions are computed for different derivatives and allows sharing of intermediate results~(see Algorithm~\ref{alg::backward_full_einsum}). Due to this sharing, we get memory savings compared to \textbf{Autodiff} approach.

{\bf TTM layer: Full Matrix}. In the Full Matrix approach, we perform the same three steps as in the Full Einsum method. The difference is that as a first contraction, we usually convolve tensors $\mathcal{X}^T$ and $\frac{\partial \mathcal{L}}{\partial \mathcal{Y}}$ along the batch axis, and the schedule of other contractions is further optimized~(see Algorithm~\ref{alg::backward_full_matrix}). It improves complexity when the product of batch size by sequence length is large (which is the case of Transformers models).

\begin{algorithm}[H]
\begin{algorithmic}[1]
\REQUIRE $\frac{\partial L}{\partial\mathcal{Y}}$; $\mathcal{X}$ 
\ENSURE $\frac{\partial L}{\partial\mathcal{X}}, \frac{\partial L}{\partial\mathcal{G}^1},\dots,\frac{\partial L}{\partial\mathcal{G}^M}$

\STATE // Results in the below for-cycle are computed only for the first batch during training and reused for others.
\FOR{$k$ in $\{1,\dots,M\}$} 
\STATE Compose $einsum_k$ expression for $\frac{\partial L}{\partial\mathcal{G}^k}$
\STATE Optimize contraction schedule for composed $einsum_k$
\ENDFOR
\FOR{$k$ in $\{1,\dots,M\}$} 
\STATE $\frac{\partial L}{\partial\mathcal{G}^k} = einsum_k(\frac{\partial L}{\partial\mathcal{Y}}, \mathcal{G}^1,\dots,\mathcal{G}^M)$
\ENDFOR
\end{algorithmic}
 \caption{Backward pass (TTM layer, Full Einsum).}
 \label{alg::backward_full_einsum}
\end{algorithm}

The memory footprints of each of these methods are in Table~\ref{tab:forward_backward}. We select the most optimal pair according to memory and time - \textbf{Einsum Forward}, \textbf{Full Matrix Backward} - and employ it in TTM layer implementation.

\begin{algorithm}[H]
\begin{algorithmic}[1]
\REQUIRE $\frac{\partial L}{\partial\mathcal{Y}}$; $\text{saved activations from forward } \mathcal{Y}_1,\dots\mathcal{Y}_M$
\ENSURE $\frac{\partial L}{\partial\mathcal{X}}, \frac{\partial L}{\partial\mathcal{G}^1},\dots,\frac{\partial L}{\partial\mathcal{G}^M}$
\STATE $\frac{\partial L}{\partial\mathcal{Y}_M} = \frac{\partial L}{\partial\mathcal{Y}}$
\FOR{$k$ in $\{M,\dots,1\}$} 
\STATE $\frac{\partial L}{\partial\mathcal{G}^k} = einsum(\mathcal{Y}_{k-1}, \frac{\partial L}{\partial\mathcal{Y}^k})$
\STATE $\frac{\partial L}{\partial\mathcal{Y}_{k-1}} = einsum(\frac{\partial L}{\partial\mathcal{Y}_k},\mathcal{G}^k)$
\ENDFOR
\STATE $\frac{\partial L}{\partial\mathcal{X}} = \frac{\partial L}{\partial\mathcal{Y}_0}$
\end{algorithmic}
 \caption{Backward pass (TTM layer, Autodiff).}
 \label{alg::backward_autodiff}
\end{algorithm}

\section{Singular Value decomposition~(SVD) Layer}

We compress the initial model by replacing fully-connected layers with their SVD analogs. 

More precisely, assuming that $W$ is a layer weight matrix, we define SVD as follows: $W = U \Sigma V^T$. Then we use truncated products of it $U_r = U[:, :r], \Sigma_r = \Sigma[:r, :r], V_r = V[:, :r]$ to define weights for two sequential linear layers, with which we will replace the current:

\begin{equation}
    W_2 = U[:, :r] \sqrt{\Sigma_r}, \\
    W_1 = \sqrt{\Sigma_r} U^T[:, :r]
\end{equation} 

As a result, we get an approximation of linear matrix $W \approx W_2 W_1$ and an approximation of the initial layer $Y \approx X W_1^T W_2^T + b$.

If $W$ have $n_{in}, n_{out}$ shape, the number of parameters in the layer before compression is $n_{in} \times n_{out}$, after representation by truncated SVD, number of parameter in layer is $r \times (n_{in} + n_{out})$.

\section{Experiments: end-to-end training of GPT-2 with custom layers}

We conducted experiments with a generative model of GPT-2.
We replaced the fully connected layers with the sequence of corresponding TTM containers and trained the resulting models from scratch on the task of language modelling~(LM). In this section, we examine the performance of the original model with our model and a GPT-2 with a fully-connected layer, replaced with SVD structure~(with the same parameter budget as our model). 

The general intuition of TTM layers superiority w.r.t. SVD is as follows: TTM is proved to be full-rank~\cite{Khrulkov}, since the truncated SVD is a low-rank method. Training the layers from scratch, we find a structure which defines weight matrices. The matrix $\mathcal{M}\in \mathbf{R}^{IJ}$ being restored from TTM containers has rank $R_{TTM} = min(I, J)$, otherwise matrix assembled from SVD factors has truncated rank $R_{SVD} < min(I, J)$

We can suggest that for matrices with a certain dimension:
\begin{itemize}
\item TTM is seeking a proper weight in a more comprehensive space by utilizing a set of full-rank matrices, which are more effective than a set of matrices with truncated ranks;
\item Higher rank matrix can store more information than a matrix with the same dimensions but a lower rank.
\end{itemize}


\begin{table*}[h!]
\small
\centering
\begin{tabular}{l | c|  c | c | c | c | c | c | c | c | c | c }
\toprule
Model & \% full GPT &  STSB & CoLA & MNLI & MRCP & QNLI & QQP & RTE & SST2 & WNLI & AVG\\
\midrule
GPT-2 med & 100 &  0.76 & 0.45 & 0.82 & 0.78 & 0.87 & 0.87 & 0.53 & 0.92 &0.43 & 0.74 \\
GPT-2 TTM-72 & 61 &  0.77 & 0.23 & 0.79 & 0.80 & 0.61 & 0.86 & 0.47 & 0.82 & 0.56 & 0.66 \\ 
GPT-2 SVD-50 & 62 & 0.73 & 0.08 & 0.78 & 0.68 & 0.84 & 0.84 & 0.57 & 0.89 & 0.43 & 0.64\\ 
DistilGPT & 23 & 0.18 & 0.00 & 0.73 & 0.70 & 0.79 & 0.52 & 0.57 & 0.88 & 0.43 & 0.53\\ 
\bottomrule
\end{tabular}
\caption{Performance for GPT-2-based model on GLUE benchmark after one epoch fine-tining.}
\label{table_glue}
\end{table*}

\begin{table}
\centering

\scalebox{0.98}{
\begin{tabular}{l | c | c | c}
\toprule
Model & ROUGE-1 & ROUGE-2 & ROUGE-L \\
\midrule
GPT-2 med &  20.5 &  4.6 & 10.2  \\ 
GPT-2 SVD-50  &  18.1 &  2.3 & 11.3 \\
GPT-2 TTM-72  &  20.1 &  4.1 & 9.9 \\
\bottomrule
\end{tabular}
}

\caption{Text summarization results.}
\label{summary}
\end{table}

\subsection{Hyperparameter selection}
The proposed layer structure assumes two sets of hyperparameters - \textit{TTM cores shapes} and \textit{TTM ranks}. 
The matrix of sizes $(I, J)$ is represented in cores $\mathcal{G}\in \mathbb{R}^{1, j_1, i_1, R_1}, \mathcal{G}\in \mathbb{R}^{R_1, j_2, i_2, R_2},\dots, \mathcal{G}\in \mathbb{R}^{R_{M-1},  j_M, i_M, 1}$, where $I = \prod_{k = 1}^M {i_k}, J = \prod_{k = 1}^M {j_k}$, M - number of cores. Assuming the formula for compression rate in a TTM layer: $$c\_rate = \frac{R(i_1 j_1 + i_m j_m) + R^2\sum_{m=2}^{M-1} i_m j_m}{\prod_{m=1^M}i_m j_m}$$, we state that for the maximum compression rate shapes should be as close to each other as possible.
We choose $i_k*j_k$ in a way that they are equal to each other and approximately equal to $(I*J)^{1/M}$. 
Shapes selection is implemented with a custom algorithm which will be presented in the source code.
In our case, GPT-2 small fully-connected layers [I, J] is [768, 3072]; $768 = 4*6*8*4$ and $3072 = 8*8*6*8$; $i_k*j_k$ are 8*4, 8*6, 6*8, 8*4.

As for the choice of ranks, we choose them based on the desired compression of the entire model. For a small GPT, these are from 50\% to 90\%. For a medium GPT, the reduction is 40\%.

\subsection{In-domain language modelling task }

To evaluate in-domain performance on the LM task, we provide training and evaluation on the train and test partition of the same dataset, respectively. We replace the fully connected layers of GPT-2-small with TTM of ranks 16, 32, 64, and 80. We train and validate the model with block size 512 on the Wikitext-103 dataset~\cite{DBLP:journals/corr/MerityXBS16} for 40 epochs using the AdamW optimizer and the Cosine warmup scheduler, increasing the training step from $0$ to $2.5e^{-4}$. In this and subsequent experiments, we established the maximum learning rate point relative to the total number of training steps. Our goal was to ensure that the model reached its highest point and underwent approximately 1/10 of the entire learning process. Table~\ref{table_gpt_small_en} shows that the resulting perplexity is comparable to the original model. However, model compression has a negligible impact on quality within this domain. For example, a reduction in parameters of over 30\% only results in a half a percent decrease in perplexity, while a reduction of over 40\% leads to a 3\% drop.

\subsection{Out-domain language modelling task}


In this setup, we perform validation on the test section of Wikitext-103 while training the model on other datasets for the same language modeling task.

We train GPT-TT architecture on a sufficiently large dataset OpenWebText~\cite{Gokaslan2019OpenWeb}, which imitates the WebText dataset and is publicly available. We train the model for 10 epoch with a similar optimizer scheduler with a maximum learning rate $2.95\exp{-5}$ and global batch size 340. Upon reaching the perplexities value of 50, we halved the batch size. We use an optimizer and scheduler as in the previous section, sequence length 1024. The optimal parameters were chosen based on the perplexity on the validation part of the Wikitext-103 dataset of a small GPT-2 model with classical fully connected layers. After obtaining the optimal parameters for the classical model, the learning settings were fixed. 
The training process continued for approximately 20 days on 4 GPUs 3090ti. To receive a GPT-based model with a compatible size, we train from scratch under the same condition the GPT-2 medium with linear layer replaced with SVD-structure layers with rank 50. As shown in Table~\ref{table_gpt_medium_en}, the best perplexity among the compressed models pertains to  OPT~\cite{Opt} with 350 million of parameters. Herewith, OPT saves 7\% of full GPT-2, while TTM-72 saves 40\%, and perplexity is decreased to 31. At the same time, an SVD-50 of a similar size as TTM-72 has perplexity 55, which is even worse than Distill GPT, the architecture with the smallest number of parameters.
 




\subsection{Natural Languge Understanding - GLUE}
We take a pre-trained GPT TTM-72 model from the previous section~(without fine-tuning) and validate it on a General Language Understanding Evaluation~(GLUE) benchmark. It is a collection of nine natural language tasks, including language acceptability, sentiment analysis, paraphrasing and natural language inference. The evaluating script is based on the original Transformer repository~\cite{wolf-etal-2020-transformers}. We add a top head compatible with the given task and run one training epoch. We choose just one epoch to avoid a situation where several models, all "large" concerning the number of tokens in the dataset but of different sizes relative to each other, converge to approximately the same loss during the entire training cycle~\cite {scaling_laws}. We repeated these experiments 5 times with different random seeds, Table~\ref{table_glue} shows the averaged obtained results with a standard deviation of no more than $0.0008$. The classical models and models with TTM layers show approximately equal results, periodically overtaking each other. GPT-2 TTM-72 has a performance decrease in Acceptability and several Question-Answering data~(QNLI, MNLI). The result of SVD-50 is close to TTM-72. 

\subsection{Text summarization }
We also compare the behaviour of proposed models on the text summarization task when tuning on a small amount of data. Based on the pipeline from~\cite{summarization}, we trained both models on 3000 objects from the CNN/Daily Mail datasets~\cite{NIPS2015_afdec700, nallapati-etal-2016-abstractive}. The obtained ROUGEs are not high~(Table~\ref{summary}) but match the result from cited paper and highlight the similar behaviour of the classical GPT-2 and TTM-72. SVD-50 shows a bit worse outcome, except for the metric ROUGE-L.

\section{Conclusion}
We introduce custom TTM layers representing fully-connected layers in a Tensor Train Matrix format. We employ this layer in a transformer-based GPT-2 architecture and obtain a 40\% lighter model, which performs in-domain, out-of-domain and downstream tasks without the significant quality drop. In addition, the GPT-2 TTM can replace GPT-2 in memory-restricted environments; TTM layers can be used to reduce the effective size of any Transformer architecture. We have also demonstrated experimentally that when learning from scratch under the same conditions, structures with lower ranks~(like SVD) are less expressive than Tensor Train Models (TTMs).


\section{Limitations}
The main limitation of this work is that a model with custom layers must be trained from scratch. It requires the operation of several industrial GPUs for several weeks and the necessary equipment - at least a load-bearing power supply. Such resources may be limited in the academy. The proposed model also wasn't be validated on few-shot tasks, which defines a good generalization ability of the pre-trained model.
It is important to recognize that training a large model from scratch is a skill that requires a certain level of expertise. As a result, the performance of two identical architectures can vary significantly depending on the specific training pipeline utilized.



\end{document}